\documentclass[graybox]{svmult}

\usepackage{type1cm}        
\usepackage{makeidx}         
\usepackage{graphicx}        
\usepackage{multicol}        
\usepackage[bottom]{footmisc}

\usepackage{newtxtext}       %
\usepackage{newtxmath}       

\makeindex             

\usepackage{amsmath}
\usepackage{amsfonts}
\usepackage{color}
\usepackage{authblk}
\usepackage{hyperref}
\usepackage{algorithm}
\usepackage{algpseudocode}
\usepackage{multirow}
\usepackage{enumitem}
\usepackage{breakcites}
\usepackage{marvosym}
\usepackage{natbib}

\newcommand{\concept}[1]{\textsc{#1}}

\begin{document}

\title*{Generalizing Psychological Similarity Spaces to Unseen Stimuli\thanks{The research presented in this paper is an updated, corrected, and significantly extended version of research reported in \cite{Bechberger2018AIC}.}}
\subtitle{Combining Multidimensional Scaling with Artificial Neural Networks}
\author{Lucas Bechberger and Kai-Uwe K\"uhnberger}
\institute{Lucas Bechberger \Letter (0000-0002-1962-1777) \at Institute of Cognitive Science, Osnabr\"uck University \email{lucas.bechberger@uni-osnabrueck.de}
\and Kai-Uwe K\"uhnberger \at Institute of Cognitive Science, Osnabr\"uck University \email{kai-uwe.kuehnberger@uni-osnabrueck.de}}
%
%
\maketitle

\abstract{The cognitive framework of conceptual spaces proposes to represent concepts as regions in psychological similarity spaces. These similarity spaces are typically obtained through multidimensional scaling (MDS), which converts human dissimilarity ratings for a fixed set of stimuli into a spatial representation. One can distinguish metric MDS (which assumes that the dissimilarity ratings are interval or ratio scaled) from nonmetric MDS (which only assumes an ordinal scale). In our first study, we show that despite its additional assumptions, metric MDS does not necessarily yield better solutions than nonmetric MDS. In this chapter, we furthermore propose to learn a mapping from raw stimuli into the similarity space using artificial neural networks (ANNs) in order to generalize the similarity space to unseen inputs. In our second study, we show that a linear regression from the activation vectors of a convolutional ANN to similarity spaces obtained by MDS can be successful and that the results are sensitive to the number of dimensions of the similarity space.}

\section{Introduction}
\label{Intro}

The cognitive framework of conceptual spaces \citep{Gardenfors2000} proposes a geometric representation of conceptual structures: Instances are represented as points and concepts are represented as regions in psychological similarity spaces. Based on this representation, one can explain a range of cognitive phenomena from one-shot learning to concept combination.\\

In principle, there are three ways of obtaining the dimensions of a conceptual space: If the domain of interest is well understood, one can manually define the dimensions and thus the overall similarity space. A second approach is based on machine learning algorithms for dimensionality reduction. For instance, unsupervised artificial neural networks (ANNs) such as autoencoders or self-organizing maps can be used to find a compressed representation for a given set of input stimuli. This task is typically solved by optimizing a mathematical error function which may be not satisfactory from a psychological point of view.\\

A third way of obtaining the dimensions of a conceptual space is based on dissimilarity ratings obtained from human subjects. One first elicits dissimilarity ratings for pairs of stimuli in a psychological study. The technique of ``multidimensional scaling'' (MDS) takes as an input these pair-wise dissimilarities as well as the desired number $t$ of dimensions. It then represents each stimulus as a point in an $t$-dimensional space in such a way that the distances between points in this space reflect the dissimilarities of their corresponding stimuli. \textit{Nonmetric} MDS assumes that the dissimilarities are only ordinally scaled and limits itself to representing the ordering of distances correctly. \textit{Metric} MDS on the other hand assumes an interval or ratio scale and also tries to represent the numerical values of the dissimilarities as closely as possible. We introduce multidimensional scaling in more detail in Section \ref{MDS}. Moreover, we present a study investigating the differences between similarity spaces produced by metric and nonmetric MDS in Section \ref{NOUN}.\\

One limitation of the MDS approach is that it is unable to generalize to unseen inputs: If a new stimulus arrives, it is impossible to directly map it onto a point in the similarity space without eliciting dissimilarities to already known stimuli. In Section \ref{Hybrid}, we propose to use ANNs in order to learn a mapping from raw stimuli to similarity spaces obtained via MDS. This hybrid approach combines the psychological grounding of MDS with the generalization capabilitiy of ANNs.

In order to support our proposal, we present the results of a first feasibility study in Section \ref{ML}: Here, we use the activations of a pre-trained convolutional network as features for a simple regression into the similarity spaces obtained in Section \ref{NOUN}.

Finally, Section \ref{Conclusions} summarizes the results obtained in this paper and gives an outlook on future work. Code for reproducing both of our studies can be found online at \url{https://github.com/lbechberger/LearningPsychologicalSpaces/} \citep{Bechberger2020GitHubPsy}.

\section{Multidimensional Scaling}
\label{MDS}

\subsection{Obtaining Dissimilarity Ratings}
\label{MDS:Ratings}

In order to collect similarity ratings from human participants, several different techniques can be used \citep{Goldstone1994,Hout2013,Wickelmaier2003}. They are typically grouped into \textit{direct} and \textit{indirect} methods: In direct methods, participants are fully aware that they rate, sort, or classify different stimuli according to their pairwise dissimilarities. Indirect methods on the other hand are based on secondary empirical measurments such as confusion probabilities or reaction times.\\

One of the classical direct techniques directly asks the participants for dissimilarity ratings. In this approach, all possible pairs from a set of stimuli are presented to participants (one pair at a time), and participants rate the dissimilarity of each pair on a continuous or categorical scale. 
Another direct technique is based on sorting tasks. For instance, participants might be asked to group a given set of stimuli into piles of similar items. In this case, similarity is binary -- either two items are sorted into the same pile or not. 

Perceptual confusion tasks can be used as an indirect technique for obtaining similarity ratings. For example, participants are asked to report as fast as possible whether two displayed items are the same or different. In this case, confusion probabilities and reaction times are measured in order to infer the underlying similarity relation.\\

\cite{Goldstone1994} has argued that the classical approaches for collecting similarity data are limited in various ways. Their biggest shortcoming is that explicitly testing all $\frac{N \cdot (N-1)}{2}$ stimulus pairs is quite time-consuming. An increasing number of stimuli therefore leads to the need for very long experimental sessions, which might cause fatigue effects. Moreover, in the course of such long sessions, participants might switch to a different rating strategy after some time, making the collected data less homogeneous.

In order to make the data collection process more time-efficient, \cite{Goldstone1994} has proposed the \textit{``Spatial Arrangement Method''} (SpAM). In this collection technique, multiple visual stimuli are displayed on a computer screen. In the beginning, the arrangement of these stimuli is randomized. Participants are then asked to arrange them via drag and drop in such a way that the distances between the stimuli are proportional to their dissimilarities. Once participants are satisfied with their solution, they can store the arrangement. The dissimilarity of two stimuli is then recorded as their Euclidean distance in pixels. As $N$ items can be displayed at once, each single modification by the user updates $N$ distance values at the same time which makes this procedure quite efficient. Moreover, SpAM quite naturally incorporates geometric constraints: If $A$ and $B$ are placed close together and $C$ is placed far away from $A$, then it cannot be very close to $B$.

As the dissimilarity information is recorded in the form of Euclidean distances, one might assume that the dissimilarity ratings obtained through SpAM are ratio scaled. This view is for instance held by \cite{Hout2014}. However, as participants are likely to make only a rough arrangement of the stimuli, this assumption might be too strong in practice. One can argue that it is therefore safer to only assume an ordinal scale. As far as we know, there have been no explicit investigations on this subject. We will provide an analysis of this topic in Section \ref{NOUN}.

\subsection{The Algorithms}
\label{MDS:Algorithms}

In this chapter, we follow the mathematical notation by \cite{Kruskal1964}, who gave the first thorough mathematical treatment of (nonmetric) multidimensional scaling.

One can typically distinguish two types of MDS algorithms \citep{Wickelmaier2003}, namely metric and nonmetric MDS. Metric MDS assumes that the dissimilarities are interval or ratio scaled while nonmetric MDS only assumes an ordinal scale. 

Both variants of MDS can be formulated as an optimization problem involving the pairwise dissimilarities $\delta_{ij}$ between stimuli and the Euclidean distances $d_{ij}$ between their corresponding points in the $t$-dimensional similarity space. More specifically, MDS involves minimizing the so-called ``stress'' which measures to which extent the spatial representation violates the information from the dissimilarity matrix:
$$stress = \sqrt{\frac{\sum_{i<j} \left(d_{ij} - \hat{d}_{ij}\right)^2}{\sum_{i<j} \left(d_{ij}\right)^2}}$$

The denominator in this equation serves as a normalization factor in order to make stress invariant to the scale of the similarity space. 

In metric MDS, we use $\hat{d}_{ij} = a \cdot \delta_{ij}$ to compute stress. This means that we look for a configuration of points in the similarity space whose distances are a linear transformation of the dissimilarities.

In nonmetric MDS, on the other hand, the $\hat{d}_{ij}$ are not obtained by a \textit{linear} but by a \textit{monotone} transformation of the dissimilarities: Let us order the dissimilarities of the stimuli in ascending order: $\delta_{i_1 j_1} < \delta_{i_2 j_2} < \delta_{i_3 j_3} < \dots$.
The $\hat{d}_{ij}$ are then obtained by defining an analogous ascening order: $\hat{d}_{i_1 j_1} < \hat{d}_{i_2 j_2} < \hat{d}_{i_3 j_3} < \dots$.
Nonmetric MDS therefore only tries to reflect the \textit{ordering} of the dissimilarities in the distances while metric MDS also tries to take into account their differences and ratios.\\

There are different approaches towards optimizing the stress function, resulting in different MDS algorithms. Kruskal's original nonmetric MDS algorithm \citep{Kruskal1964a} is based on gradient descent: In an interative procedure, the derivative of the stress function with respect to the coordinates of the individual points is computed and then used to make a small adjustment to these coordinates. Once the derivative becomes close to zero, a minimum of the stress fuction has been found.

A more recent MDS algorithm by \cite{DeLeeuw1977} is called SMACOF (an acronym of ``\textbf{S}caling by \textbf{Ma}jorizing a \textbf{Co}mplicated \textbf{F}unction''). De Leeuw pointed out that Kruskal's gradient descent method has two major shortcomings: Firstly, if the points for two stimuli coincide (i.e., $x_i = x_j$), then the distance function of these two points is not differentiable. Secondly, Kruskal was not able to give a proof of convergence for his algorithm.

In order to overcome these limitations, De Leeuw showed that minimizing the stress function is equivalent to maximizing another function $\lambda$ which depends on the distances and dissimilarities. This function can be easily expressed by matrix multiplications of the configuration of points in the similarity space with two matrices. De Leeuw proved that by iteratively multiplying the configuration with these matrices, one can maximize $\lambda$ and thus minimize stress. Moreover, one can prove that this iterative procedure converges. SMACOF is computationally efficient and guarantees a monotone convergence of stress \cite[Chapter 8]{Borg2005}.\\

Picking the right number of dimensions $t$ for the similaritiy space is not trivial. \cite{Kruskal1964} proposes two approaches to address this problem: 

On the one hand, one can create a so-called ``Scree'' plot that shows the final stress value for different values of $t$. If one can identify an ``elbow'' in this diagram (i.e., a point after which the stress decreases much slower than before), this can point towards a useful value of $t$. 

On the other hand, one can take a look at the interpretability of the generated configurations. If the optimal configuration in a $t$-dimensional space has a sufficient degree of interpretability and if the optimal configuration in a $t+1$ dimensional space does not add more structure, then a $t$-dimensional space might be sufficient.

\section{Extracting Similarity Spaces from the NOUN Data Set}
\label{NOUN}

It is debatable whether metric or nonmetric MDS should be used with data collected through SpAM. Nonmetric MDS makes less assumptions about the underlying measurement scale and therefore seems to be the ``safer'' choice. If the dissimilarities are however ratio scaled, then metric MDS might be able to harness these pieces of information from the distance matrix as additional constraints. This might then result in a semantic space of higher quality.\\

In our study, we compare metric to nonmetric MDS on a data set obtained through SpAM.
If the dissimilarities obtained through SpAM are not ratio scaled, then the main assumption of metric MDS is violated. We would then expect that nonmetric MDS yields better solutions than metric MDS. If the dissimilarities obtained through SpAM are however ratio scaled and if the differences and ratios of dissimilarities do contain considerable amounts of additional information, then metric MDS should have a clear advantage over nonmetric MDS.

\begin{figure}[t]
	\centering
	\includegraphics[width = \columnwidth]{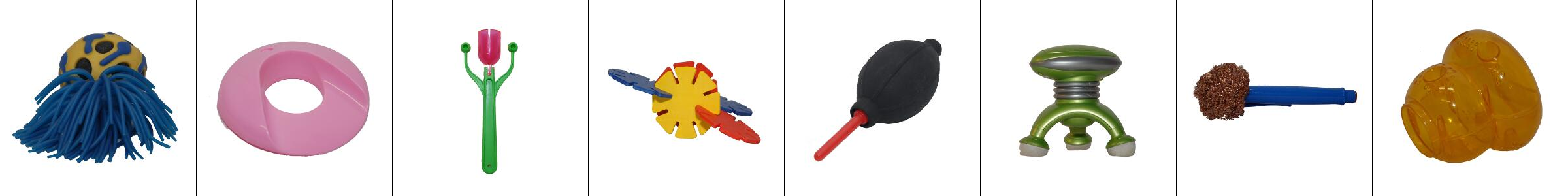} 
	\caption{Eight example stimuli from the NOUN data set \citep{Horst2016}.}
	\label{MDS:fig:NOUN_Examples}
\end{figure}

For our study, we used existing dissimilarity ratings reported for the Novel Object and Unusual Name (NOUN) data set \citep{Horst2016}, a set of 64 images of three-dimensional objects that are designed to be novel but also look naturalistic. Figure \ref{MDS:fig:NOUN_Examples} shows some example stimuli from this data set.

\subsection{Evaluation Metrics}
\label{MDS:NOUN:Evaluation}

We used the \texttt{stress0} function from R's \texttt{smacof} package to compute both metric and nonmetric stress. We expect stress to decrease as the number of dimensions increases. 
If the data obtained through SpAM is ratio scaled, then we would expect that metric MDS achieves better values on metric stress (and potentially also on nonmetric stress) than nonmetric MDS. If the SpAM dissimilarities are not ratio scaled, then metric MDS should not have any advantage over nonmetric MDS.\\

Another possible way of judging the quality of an MDS solution is to look for interpretable directions in the resulting space. However, \cite{Horst2016} have argued that for the novel stimuli in their data set there are no obvious directions that one would expect. Without a list of candidate directions, an efficient and objective evaluation based on interpretable directions is however hard to achieve. We therefore did not pursue this way of evaluating similarity spaces.\\

As an additional way of evaluation, we measured the correlation between the distances in the MDS space and the dissimilarity scores from the psychological study.

\textbf{Pearson's $r$} \citep{Pearson1895} measures the linear correlation of two random variables by dividing their covariance by the product of their individual variances. Given two vectors $x$ and $y$ (each containing $N$ samples from the random variables $X$ and $Y$, respectively), Pearson's $r$ can be estimated as follows, where $\bar{x}$ and $\bar{y}$ are the average values of the two vectors: 
$$r_{xy} = \frac{\sum_{i=1}^n (x_i - \bar{x})(y_i - \bar{y})}{\sqrt{\sum_{i=1}^n (x_i - \bar{x})^2} \sqrt{\sum_{i=1}^n (y_i - \bar{y})^2}}$$

\textbf{Spearman's $\rho$} \citep{Spearman1904} generalizes Pearson's $r$ by allowing also for nonlinear monotone relationships between the two variables. It can be computed by replacing each observation $x_i$ and $y_i$ with its corresponding rank, i.e., its index in a sorted list, and by then computing Pearson's $r$ on these ranks. By replacing the actual values with their ranks, the numeric distances between the sample values lose their importance -- only the correct ordering of the samples remains important. Like Pearson's $r$, Spearman's $\rho$ is confined to the interval [-1, 1] with positive values indicating a monotonically increasing relationship.

Both MDS variants can be expected to find a configuration such that there is a monotone relationship between the distances in the similarity space and the original dissimilarity matrix. That is, smaller dissimilarites correspond to smaller distances and larger dissimilarities correspond to larger distances. For Spearman's $\rho$, we therefore do not expect any notable differences between metric and nonmetric MDS. For metric MDS, we also expect there to be a \emph{linear} relationship between dissimilarities and distances. Therefore, if the dissimilarities obtained by SpAM are ratio scaled, then metric MDS should give better results with respect to Pearson's $r$ than nonmetric MDS. \\

A final way for evaluating the similarity spaces obtained by MDS is visual inspection: If a visualization of a given similarity space shows meaningful structures and clusters, this indicates a high quality of the semantic space. We limit our visual inspection to two-dimensional spaces.

\subsection{Methods}
\label{MDS:NOUN:Methods}

In order to investigate the differences between metric and nonmetric MDS in the context of SpAM, we used the SMACOF algorithm in its original implementation in R's \texttt{smacof} library.\footnote{See \url{https://cran.r-project.org/web/packages/smacof/smacof.pdf}} SMACOF can be used in both a \emph{metric} and a \emph{nonmetric} variant. The underlying algorithm stays the same, only the definition of stress and thus the optimization target differs. Both variants were explored in our study. We used 256 random starts with the maximum number of iterations per random start set to 1000. The overall best result over these 256 random starts was kept as final result.

For each of the two MDS variants, we constructed MDS spaces of different dimensionalities (ranging from one to ten dimensions). For each of these resulting similarity spaces, we computed both its metric and its nonmetric stress.\\

In order to analyze how much information about the dissimilarities can be readily extracted from the images of the stimuli, we also introduced two baselines.

For our first baseline, we used the similarity of downscaled images: For each original image (with both a width and height of 300 pixels), we created lower-resolution variants by aggregating all the pixels in a $k \times k$ block into a single pixel (with $k \in [2,300]$). We compared different aggregation functions, namely, minimum, mean, median, and maximum. The pixels of the resulting downscaled image were then interpreted as a point in a $\lceil \frac{300}{k} \rceil \times \lceil \frac{300}{k} \rceil$ dimensional space.

For our second baseline, we extracted the activation vectors from the second-to-last layer of the Inception-v3 network \citep{Szegedy2016} for each of the images from the NOUN data set. Each stimulus was thus represented by its corresponding activation pattern. While the downscaled images represent surface level information, the activation patterns of the neural network can be seen as more abstract representation of the image.

\begin{figure}[t]
	\centering
	\includegraphics[width = 0.7\columnwidth]{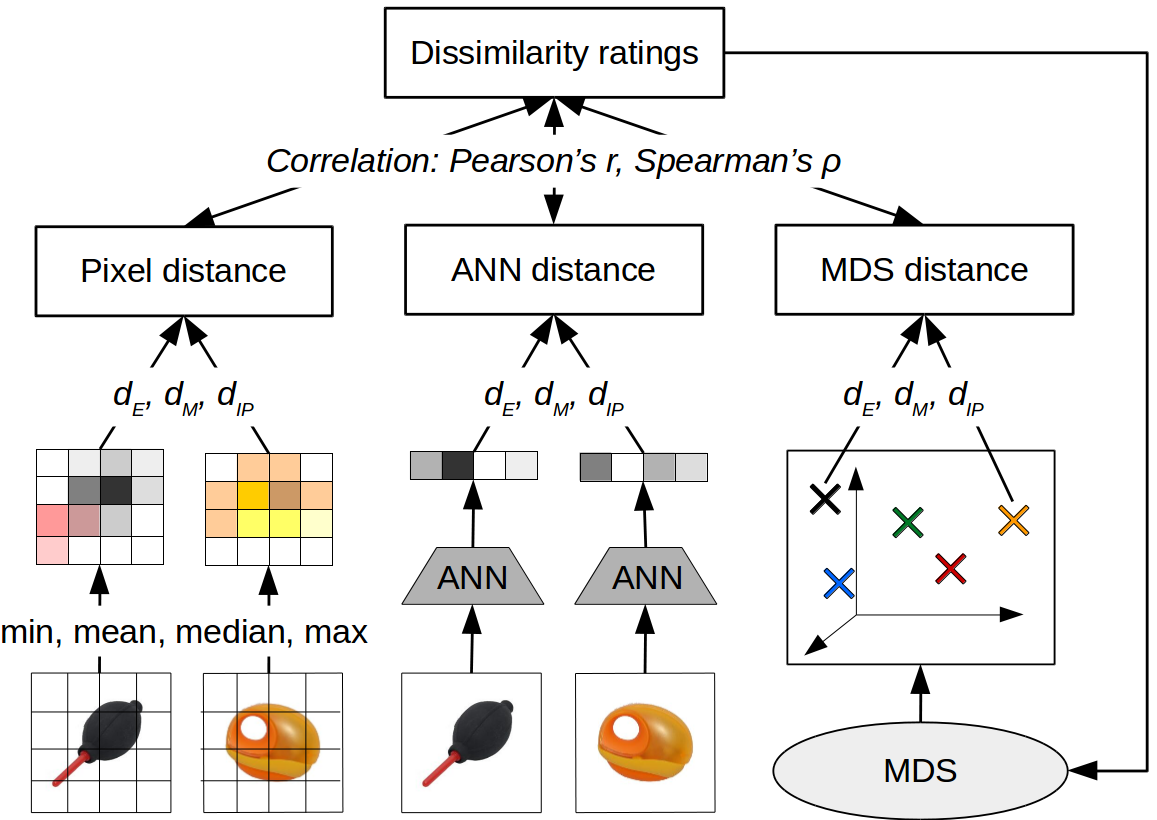} 
	\caption{Illustration of our analysis setup: We measure the correlation between the dissimilarity ratings and distances from three different sources: The pixels of downscaled images (left), activations of an artificial neural network (middle) and similarity spaces obtained by MDS (right).}
	\label{MDS:fig:NOUN_Analysis_Procedure}
\end{figure}

For each of the three representation variants (downscaled images, ANN activations, and points in an MDS-based similarity space), we computed three types of distances between all pairs of stimuli: The Euclidean distance $d_E$, the Manhattan distance $d_M$, and the negated inner product $d_{IP}$. We only report results for the best choice of the distance function. For each distance function, we used two variants: One where all dimensions are weighted equally and another one where optimal weights for the individual dimensions were estimated based on a non-negative least squares regression in a five-fold cross validation (cf. \cite{Peterson2018} who followed a similar procedure).
For each of the resulting distance matrices, we compute the two correlation coefficients with respect to the target dissimilarity ratings. We consider only matrix entries above the diagonal as the matrices are symmetric and as all entries on the diagonal are guaranteed to be zero. Our overall workflow is illustrated in Figure \ref{MDS:fig:NOUN_Analysis_Procedure}.

\subsection{Results}
\label{MDS:NOUN:Results}

\begin{figure}[t]
	\centering
	\includegraphics[width = 0.8\columnwidth]{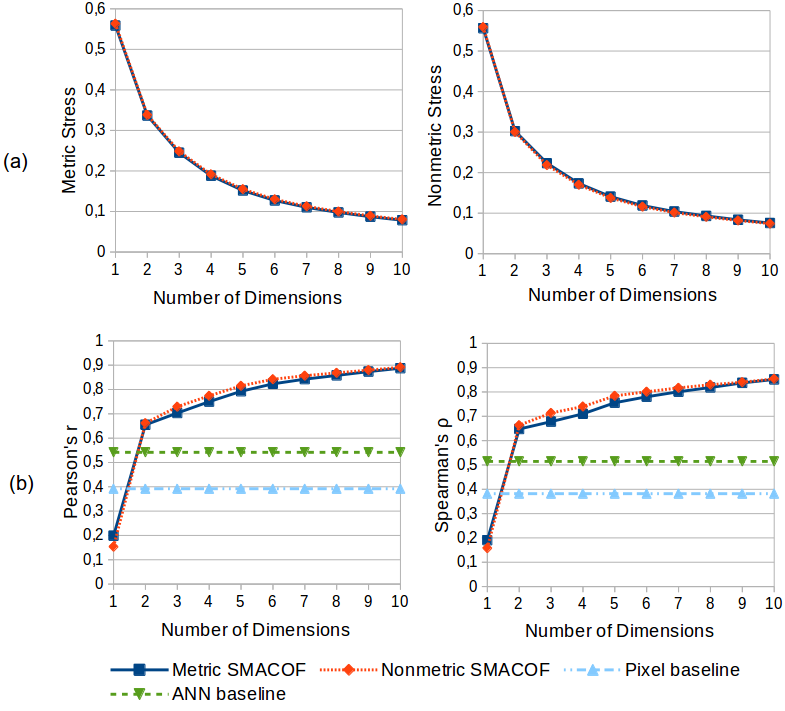}
	\caption{(a) Scree plots for both metric and nonmetric stress. (b) Correlation evaluation for the different MDS solutions and the two baselines.}
	\label{fig:NOUN_Plots}
\end{figure}

Figure \ref{fig:NOUN_Plots}a shows the Scree plots of the two MDS variants for both metric and nonmetric stress. As one would expect, stress decreases with an increasing number of dimensions: More dimensions help to represent the dissimilarity ratings more accurately.
Metric and nonmetric SMACOF yield almost identical performance with respect to both metric and nonmetric stress. This suggests that interpreting the SpAM dissimilarity ratings as ratio scaled is neither helpful nor harmful.\\

Figure \ref{fig:NOUN_Plots}b shows some line diagrams illustrating the results of the correlation analysis for the MDS-based similarity spaces. 
For both baselines, the usage of optimized weights considerably improved performance. As we can see, the ANN baseline outperforms the pixel baseline with respect to both evalulation metrics, indicating that raw pixel information is less useful in our scenario than the more high-level features extracted by the ANN. For the pixel baseline, we observed that the minimum aggregator yielded the best results.

We also observe in Figure \ref{fig:NOUN_Plots}b that the MDS solutions provide us with a better reflection of the dissimilarity ratings than both pixel-based and ANN-based distances if the similarity space has at least two dimensions. This comes as no surprise as the MDS solutions are directly based on the dissimilarity ratings whereas both baselines do not have access to the dissimilarity information. It therefore seems like our naive image-based ways of defining dissimilarities are not sufficient.

With respect to the different MDS variants, also the correlation analysis confirms our observations from the Scree plots: Metric and nonmetric SMACOF are almost indistinguishable. This again supports the view that the assumption of ratio scaled dissimilarity ratings is neither beneficial nor harmful for out data set. Moreover, we find the tendency of improved performance with an increasing number of dimensions. This again illustrates that MDS is able to fit more information into the space if this space has a larger dimensionality.\\

\begin{figure}[t]
	\centering
	\includegraphics[width = 0.48\columnwidth]{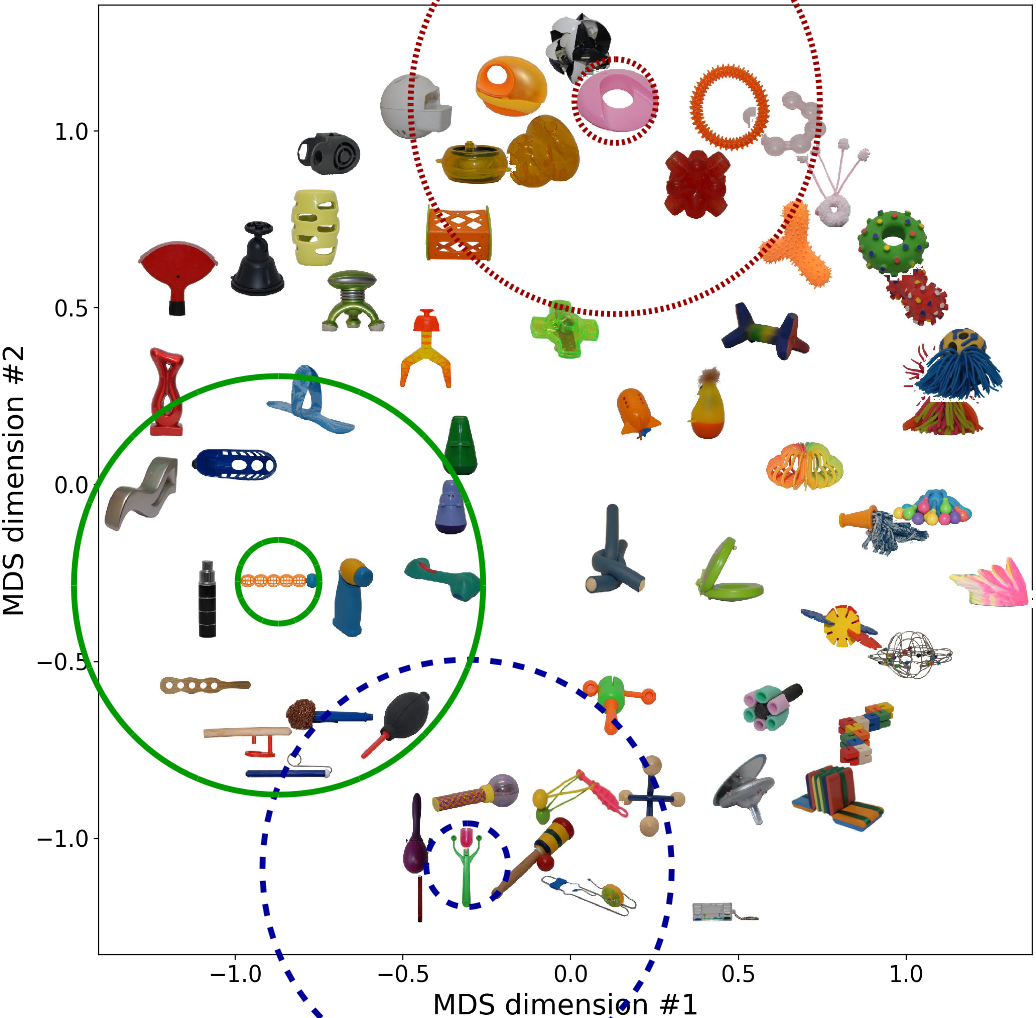} \hspace{0.2cm}
	\includegraphics[width = 0.48\columnwidth]{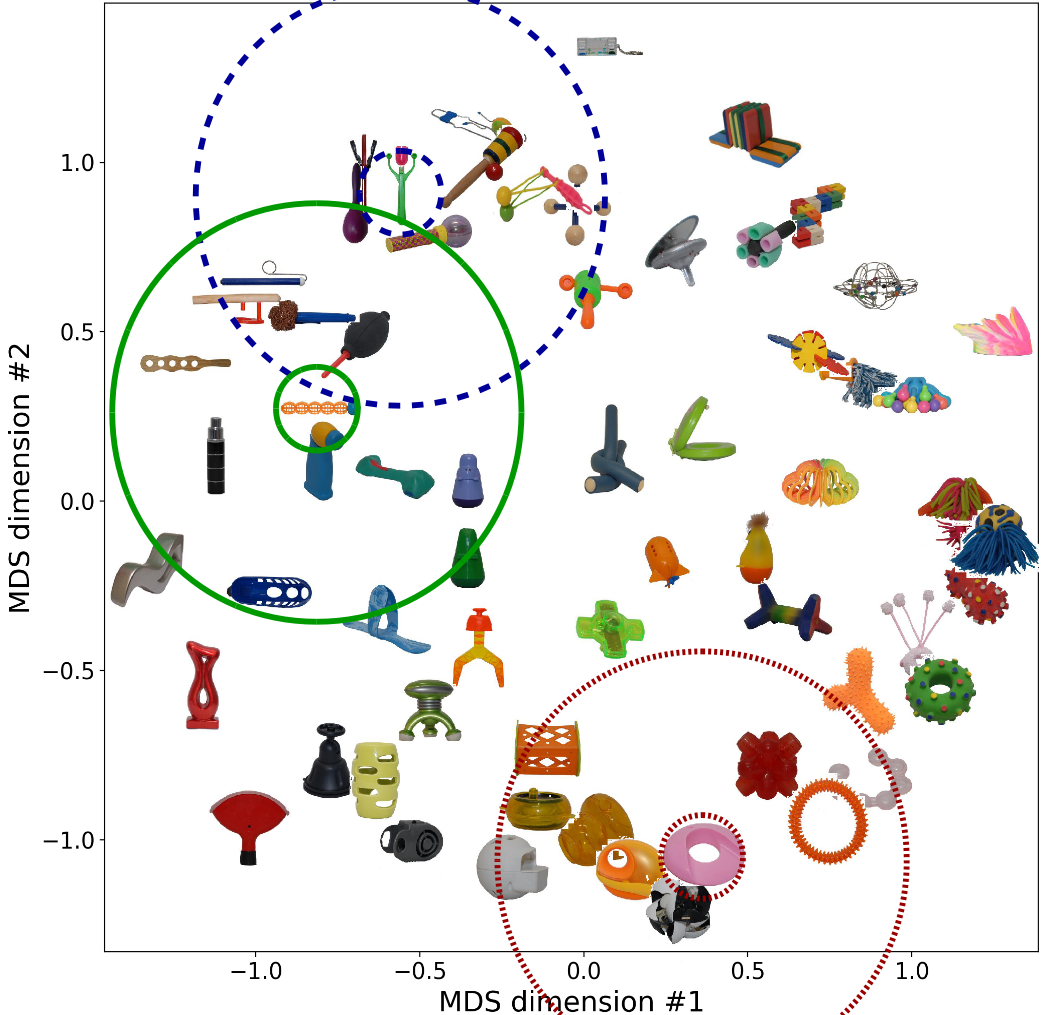}
	\caption{Illustration of the two-dimensional spaces obtained by metric SMACOF (left) and nonmetric SMACOF (right).}
	\label{fig:NOUN_2D_Spaces}
\end{figure}

Finally, let us look at the two-dimensional spaces generated by the different MDS algorithms in order to get an intuitive feeling for their semantic structure. Figure \ref{fig:NOUN_2D_Spaces} shows these spaces along with the local neighborhood of three selected items. These neighborhoods illustrate that in both spaces stimuli are grouped in a meaningful way. From our visual inspection, it seems that both MDS variants result in comparable semantic spaces with a similar structure.\\

Overall, we did not find any systematic difference between metric and nonmetric MDS on the given data set. It thus seems that the metric assumption is neither beneficial nor harmful when trying to extract a similarity space. On the one hand, we cannot conclude that the dissimilarites obtained through SpAM are \textit{not} ratio scaled. On the other hand, the additional information conveyed by differences and ratios of dissimilarities does not seem to impact the overall results. We therefore advocate the usage of nonmetric MDS due to the smaller amount of assumptions made about the dissimilarity ratings.

\section{A Hybrid Approach}
\label{Hybrid}

\subsection{Our Proposal}
\label{Hybrid:Proposal}

Multidimensional scaling (MDS) is directly based on human similarity ratings and leads therefore to conceptual spaces which can be considered psychologically valid. The prohibitively large effort required to elicit such similarity ratings on a large scale however confines this approach to a small set of fixed stimuli. We propose to use machine learning methods in order to generalize the similarity spaces obtained by MDS to unseen stimuli.
More specifically, we propose to use MDS on human similarity ratings to ``initialize'' the similarity space and artificial neural networks (ANNs) to learn a mapping from stimuli into this similarity space. 

\begin{figure}[t]
	\centering
	\includegraphics[width = 1.0\columnwidth]{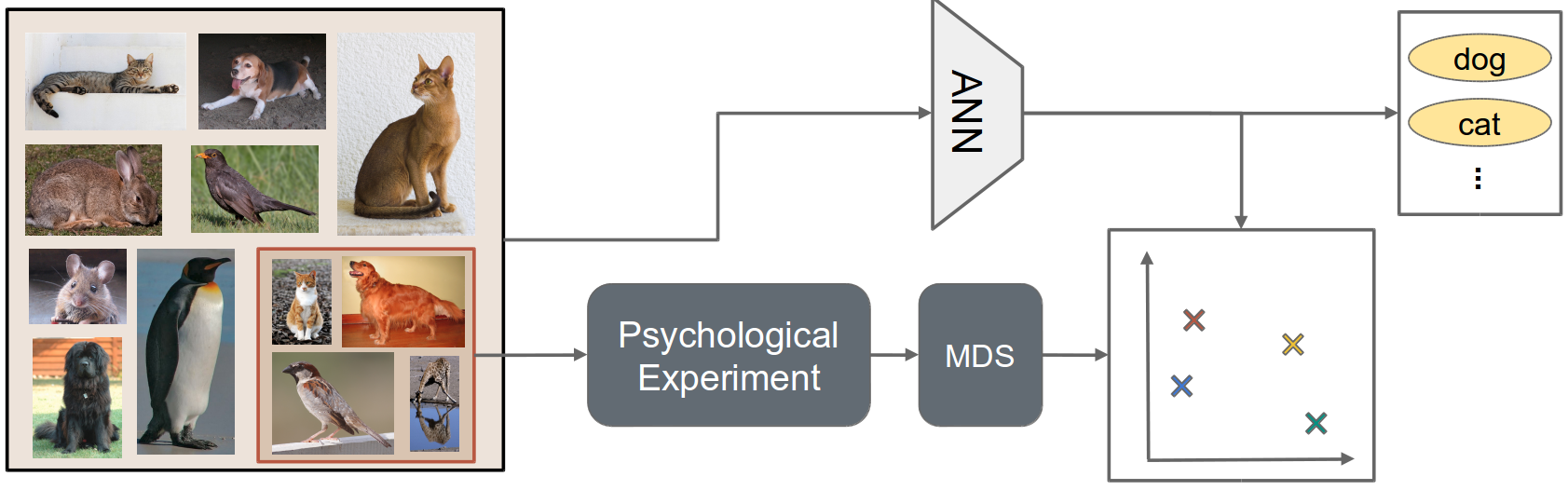} 
	\caption{Illustration of the proposed hybrid procedure: A subset of data is used to construct a conceptual space via MDS. A neural network is then trained to map images into this similarity space, aided by a secondary task (e.g., classification).}
	\label{Hybrid:fig:HybridProcedure}
\end{figure}

In order to obtain a solution having both the psychological validity of MDS spaces and the possibility to generalize to unseen inputs as typically observed for neural networks, we propose the following hybrid approach, which is illustrated in Figure \ref{Hybrid:fig:HybridProcedure}:\\

After having determined the domain of interest (e.g., the domain of animals), one first needs to acquire a data set of stimuli from this domain. This data set should cover a wide variety of stimuli and it should be large enough for applying machine learning algorithms. Using the whole data set with potentially thousands of stimuli in a psychological experiment is however unfeasible in practice. Therefore, a relatively small, but still sufficiently representative subset of these stimuli needs to be selected for the elicitation of human dissimilarity ratings. 
This subset of stimuli is then used in a psychological experiment where dissimilarity judgments by humans are obtained, using one of the techniques described in Section \ref{MDS:Ratings}.

In the next step, one can apply MDS to the collected dissimilarity judgments in order to extract a spatial representation of the underlying domain. As stated in Section \ref{MDS:Algorithms}, one needs to manually select the desired number of dimensions -- either based on prior knowledge or by manually optimizing the trade-off between high representational accuracy and a low number of dimensions. The resulting similarity space should ideally be analyzed for meaningful structures and a high correlation of inter-point distances to the original dissimilarity ratings. 

Once this mapping from stimuli (e.g., images of animals) to points in a similarity space has been established, we can use it in order to derive a ground truth for a machine learning problem: We can simply treat the stimulus-point mappings as labeled training instances where the stimulus is identified with the input vector and the point in the similarity space is used as its label. We can therefore set up a regression task from the stimulus space to the similarity space.\\

Artificial neural networks (ANNs) have been shown to be powerful regressors that are capable of discovering highly non-linear relationships between raw low-level stimuli (such as images) and desired output variables. They are therefore a natural choice for this task. ANNs are typically a very data-hungry machine learning method -- they need large amounts of training examples and many training iterations in order to achieve good performance. However, the available number of stimulus-point pairs in our proposed procedure is quite low for a machine learning problem -- as argued before, we can only look at a small number of stimuli in a psychological experiment. 

We propose to resolve this dilemma not only through data augmentation, but also by introducing an additional training objective (e.g., correctly classifying the given images into their respective classes such as \concept{cat} and \concept{dog}). This additional training objective can also be optimized on all the remaining stimuli from the data set that have not been used in the psychological experiment. Using a secondary task with additional training data constrains the network's weights and can be seen as a form of regularization: These additional constraints are expected to counteract overfitting tendencies, i.e., tendencies to memorize all given mapping examples without being able to generalize.\\

Figure \ref{Hybrid:fig:HybridProcedure} illustrates the secondary task of predicting the correct classes. This approach is only applicable if the data set contains class labels. 
If the network is forced to learn a classification task, then it will likely develop an internal representation where all members of the same class are represented in a similar way. The network then ``only'' needs to learn a mapping from this internal representation (which presumedly already encodes at least some aspects of a similarity relation between stimuli) into the target similarity space. 

Another secondary task consists in reconstructing the original images from a low-dimensional internal representation, using the structure of an autoencoder. As the computation of the reconstruction error does not need any class labels, this is applicable also to unlabeled data sets, which are in general larger and easier obtain than labeled data sets. 
The network needs to accurately reconstruct the given stimuli while using only information from a small bottleneck layer. The small size of the bottleneck layer creates an incentive to encode similar input stimuli in similar ways such that the corresponding reconstructions are also similar to each other. Again, this similarity relation learned from the overall data set might be useful for learning the mapping into the similarity space. The autoencoder structure has the additional advantage that one can use the decoder network to generate an image based on a point in the conceptual space. This can be a useful tool for visualization and further analysis.\\

One should be aware that there is a difference between perceptual and conceptual similarity: Perceptual similarity focuses on the similarity of the raw stimuli, e.g., with respect to their shape, size, and color. Conceptual similarity on the other hand takes place on a more abstract level and involves conceptual information such as the typical usage of an object or typical locations where a given object might be found. For instance, a violin and a piano are perceptually not very similar as they have different sizes and shapes. Conceptually, they might be however quite similar as they are both musical instruments that can be found in an orchestra.

While class labels can be assigned on both the perceptual (\concept{round} vs. \concept{elongated}) and the conceptual level (\concept{musical instrument} vs. \concept{fruit}), the reconstruction objective always operates on the perceptual level. If the similarity data collected in the psychological experiment is of perceputal nature, then both secondary tasks seem promising. If we however target conceptual similarity, then the classification objective seems to be the preferable choice.

\subsection{Related Work}
\label{Hybrid:RelatedWork}

\cite{Peterson2017, Peterson2018} have investigated whether the activation vectors of a neural network can be used to predict human similarity ratings. They argue that this can enable researchers to validate psychological theories on large data sets of real world images. 

In their study, they used six data sets containing 120 images (each 300 by 300 pixels) of one visual domain (namely, animals, automobiles, fruits, furniture, vegetables, and ``various''). Peterson et al. conducted a psychological study which elicited pairwise similarity ratings for all pairs of images using a Likert scale. When applying multidimensional scaling to the resulting dissimilarity matrix, they were able to identify clear clusters in the resulting space (e.g., all birds being located in a similar region of the animal space). Also when applying a hierarchical clustering algorithm on the collected similarity data, a meaningful dendrogram emerged.

In order to extract similarity ratings from five different neural networks, they computed for each image the activation in the second-to-last layer of the network. Then for each pair of images, they defined their similarity as the inner product ($u^Tv = \sum_{i=1}^n u_i v_i$) of these activation vectors. When applying MDS to the resulting dissimilarity matrix, no meaningful clusters were observed. Also a hierarchical clustering did not result in a meaningful dendrogram. When considering the correlation between the dissimilarity ratings obtained from the neural networks and the human dissimilarity matrix, they were able to achieve values of $R^2$ between 0.19 and 0.58 (depending on the visual domain).

Peterson et al. found that their results considerably improved when using a weighted version of the inner product ($\sum_{i=1}^n w_i u_i v_i$): Both the similarity space obtained by MDS and the dendrogram obtained by hierarchical clustering became more human-like. Moreover, the correlation between the predicted similarities and the human similarity ratings increased to values of $R^2$ between 0.35 and 0.74.

While the approach by Peterson et al. illustrates that there is a connection between the features learned by neural networks and human similarity ratings, it differs from our proposed approach in one important aspect: Their primary goal is to find a way to predict the similarity ratings directly. Our research on the other hand is focused on predicting points in the underlying similarity space.\\

\cite{Sanders2018} have used a data set containing 360 pictures of rocks along with an eight-dimensional similarity space for a study which is quite similar in spirit to what we will present in Section \ref{ML}. Their goal was to train an ensemble of convolutional neural networks for predicting the correct coordinates in the similarity space for each rock image from the data set. As the data set is considerably too small for training an ANN from scratch, they used a pre-trained network as a starting point. They removed the topmost layers and replaced them by untrained fully connected layers with an output of eight linear units, one per dimension of the similarity space. In order to increase the size of their data set, they applied data augmentation methods by flipping, rotating, cropping, stretching and shrinking the original images. 

Their results on the test set showed a value of $R^2$ of 0.808, which means that over 80 percent of the variance was accounted for by the neural network. Moreover, an exemplar model on the space learned by the convolutional neural network was able to explain 98.9 percent of the variance seen in human categorization performance.

The work by Sanders and Nosofsky is quite similar in spirit to our own approach: Like we, they train a neural network to learn the mapping between images and a similarity space extracted from human similarity ratings. They do so by resorting to a pretrained neural network and by using data augmentation techniques. While they use a data set of 360 images, we are limited to an even smaller data set containing only 64 images. This makes the machine learning problem even more challanging. Moreover, the data set used by Sanders and Nosofky is based on real objects, whereas our study investigates a data set of novel and unknown objects. Finally, while they confine themselves to a single target similarity space for their regression task, we investigate the influence of the target space on the overall results.

\section{Machine Learning Experiments}
\label{ML}

In order to validate whether our proposed approach is worth pursuing, we conducted a feasibility study based on the similarity spaces obtained for the NOUN data set in Section \ref{NOUN}. Instead of training a neural network from scratch, we limit ourselved to a simple regression on top of a pre-trained image classification network. With the three experiments in our study, we address the following three research questions, respectively:
\begin{enumerate}
	\item Can we learn a useful mapping from coloured images into a low-dimensional psychological similarity space from a small data set of novel images for which no background knowledge is available?\\
	\textit{Our prediction: The learned mapping is able to clearly beat a simple baseline. However, it does not reach the level of generalization observed in the study of \cite{Sanders2018} due to the smaller amount of data available.}
	\item How does the MDS algorithm being used to construct the target similarity space influence the results?\\
	\textit{Our prediction: There is are no considerable differences between metric and nonmetric MDS.}
	\item How does the size of the target similarity space (i.e., the number of dimensions) influence the machine learning results?\\
	\textit{Our prediction: Very small target spaces are not able to reflect the similarity ratings very well and do not contain much meaningful structure. Very large target spaces on the other hand increase the number of parameters in the model which makes overfitting more likely. By this reasoning, medium-sized target spaces should provide a good trade-off and therefore the best regression performance.}
\end{enumerate}

\subsection{Methods}
\label{ML:Methods}

Please recall from Section \ref{NOUN} that the NOUN data base contains only 64 images with an image size of 300 by 300 pixels. As this number of training examples is too low for applying machine learning techniques, we augmented the data set by applying random crops, a Gaussian blur, additive Gaussian noise, affine transformations (i.e., rotations, shears, translations, scaling), and by manipulating the image's contrast and brightness. These augmentation steps were executed in random order and with randomized parameter settings. For each of the original 64 images, we created 1,000 augmented versions, resulting in a data set of 64,000 images in total. We assigned the target coordinates of the original image to each of the 1,000 augmented versions.\\

For our regression experiments, we used two different types of feature spaces: The pixels of downscaled images and high-level activation vectors of a pre-trained neural network.

For the ANN-based features, we used the Inception-v3 network \citep{Szegedy2016}. For each of the augmented images, we used the activations of the second-to-last layer as a 2048-dimensional feature vector. Instead of training both the mapping and the classification task simultaneously (as discussed in Section \ref{Hybrid}), we use an already pre-trained network and augment it by an additional output layer.

As a comparison to the ANN-based features, we used an approach similar to the pixel baseline from Section \ref{MDS:NOUN:Methods}: We downscaled each of the augmented images by dividing it into equal-sized blocks and by computing the minimum (which has shown the best correlation to the dissimilarity ratings in Secton \ref{MDS:NOUN:Results}) across all values in each of these blocks as one entry of the feature vector. We used block sizes of 12 and 24, resulting in feature vectors of size 1875 and 507, respectively (based on three color channels for downscaled images of size 25 x 25 and 13 x 13, respectively). By using these two pixel-based feature spaces we can analyze differences between low-dimensional and high-dimensional feature spaces. As the high-dimensional feature space is in the same order of magnitude as the ANN-based feature space, we can also make a meaningful comparison betwen pixel-based features and ANN-based features.\\

We compare our regression results to the zero baseline, which always predicts the origin of the coordinate system. In preliminary experiments, it has shown to be superior to any other simple baselines (such as e.g., drawing from a normal distribution estimated from the training targets).
We do not expect this baseline to perform well in our experiments, but it defines a lower performance bound for the regressors.\\

In our experiments, we limit ourselves to two simple off-the-shelf regressors, namely a linear regression and a lasso regression. Let $N$ be the number of data points, $t$ be the number of target dimensions, $y_d^{(i)}$ the target value of data point $i$ in dimension $d$ and $f_d^{(i)}$ the prediction of the regressor for data point $i$ in dimension $d$.

Both of our regressors make use of a simple linear model for each of the dimensions in the target space:
$$f_d = w_0^{(d)} + \sum_{k=1}^K w_k^{(d)} x_k$$
Here, $K$ is the number of features and $x$ is the feature vector. In a linear least-squares regression, the weights $w_k^{(d)}$ of this model are estimated by minimizing the mean squared error between the model's predictions and the actual ground truth value:
$$MSE_d = \frac{1}{N} \sum_{i=1}^N \left( y_d^{(i)} - f_d^{(i)}\right)^2$$

As the number of features is quite high, even a linear regression needs to estimate a large number of weights. In order to prevent overfitting, we  also consider a lasso regression which additionally incorporates the $L_1$ norm of the weight matrix as regularization term. It minimizes the following objective:
$$\frac{1}{N} \sum_{i=1}^N \left( y_d^{(i)} - f_d^{(i)}\right)^2 + \beta \cdot \frac{1}{K} \cdot \sum_{k=1}^K w_k^{(d)}$$
The first part of this objective corresponds to the mean squared error of the linear model's predictions, while the second part corresponds to the overall size of the weights. If the constant $\beta$ is tuned correctly, this can prevent overfitting and thus improve performance on the test set. In our experments, we investigated the following values:
$$\beta \in \{0.0, 0.001, 0.002, 0.005, 0.01, 0.02, 0.05, 0.1, 0.2, 0.5, 1.0, 2.0, 5.0, 10.0\}$$ Please note that $\beta = 0$ corresponds to an ordinary linear least-squares regression. \\

With our experiments, we would also like to investigate whether learning a mapping into a psychological similarity space is easier than learning a mapping into an arbitrary space of the same dimensionality. In addition to the real regression targets (which are the coordinates from the similarity space obtained by MDS), we created another set of regression targets by randomly shuffling the assignment from images to target points. We ensured that all augmented images created from the same original image were still mapped onto the same target point. With this shuffling procedure, we aimed to destroy any semantic structure inherent in the target space. We expect that the regression works better for the original targets than for the shuffled targets.\\

In order to evaluate both the regressors and the baseline, we used three different evaluation metrics. 
\begin{itemize}
	\item The \textbf{mean squared error (MSE)} sums over the average squared difference between the prediction and the ground truth for each output dimension.\\
	$$MSE = \sum_{d=1}^t \frac{1}{N} \cdot \sum_{i=1}^{N} \left( y_d^{(i)} - f_d^{(i)}\right)^2$$
	\item The \textbf{mean euclidean distance (MED)} provides us with a way of quantifying the average distance between the prediction and the target in the similarity space.\\
	$$MED =  \frac{1}{N} \cdot \sum_{i=1}^{N} \sqrt{ \sum_{d=1}^t \left( y_d^{(i)} - f_d^{(i)}\right)^2}$$
	\item The \textbf{coefficient of determination $R^2$} can be interpreted as the amount of variance in the targets that is explained by the regressor's predictions.\\
	\begin{align*}
	R^2 = \frac{1}{t} \cdot \sum_{d=1}^t \left(1 - \frac{S_{residual}^{(d)}}{S_{total}^{(d)}}\right) \text{ with } &S_{residual}^{(d)} = \sum_{i=1}^N \left(y_d^{(i)} - f_d^{(i)}\right)^2\\
	\text{ and } &S_{total}^{(d)} = \sum_{i=1}^N \left(y_d^{(i)} - \bar{y}\right)^2
	\end{align*}
\end{itemize}

We evaluated all regressors using an eight-fold cross validation approach, where each fold contains all the augmented images generated from eight of the original images. In each iteration, one of these folds was used as test set, whereas all other folds were used as training set. We aggregated all predictions over these eight iterations (ending up with exactly one prediction per data point) and computed the evaluation metrics on this set of aggregated predictions. 

\subsection{Experiment 1: Comparing Feature Spaces and Regressors}
\label{ML:Experiment_1}

In our first experiment, we want to test the following hypotheses:
\begin{enumerate}
	\item The learned mapping is able to clearly beat the baseline. However, it does not reach the level of generalization observed in the study of \cite{Sanders2018} due to the smaller amount of data available.
	\item A regression from the ANN-based features is more successful than a regression from the pixel-based features.
	\item As the similarity spaces created by MDS encode semantic similarity by geometric distance, we expect that learning the correct mapping generalizes better to the test set than learning a shuffled mapping.
	\item As the feature vectors are quite large, the linear regression has a large number of weights to optimize, inviting overfitting. Regularization through the $L_1$ loss included in the lasso regressor can help to reduce overfitting.
	\item For smaller feature vectors, we expect less overfitting tendencies than for larger feature vectors. Therefore, less regularization should be needed to achieve optimal performance.	
\end{enumerate}

Here, we limit ourselves to a single target space, namely the four-dimensional similarity space obtained by \cite{Horst2016} through metric MDS. \\

\begin{table}[t]
  \centering
  \begin{tabular}{|c|c|c||c|c|c||c|c|c||c|}
    \hline
    	\multirow{2}{*}{\textbf{Regression}}	& \multirow{2}{*}{\textbf{Feature Space}} & \multirow{2}{*}{\textbf{Targets}}	& \multicolumn{3}{|c||}{\textbf{Test Set Performance}} & \multicolumn{3}{|c||}{\textbf{Degree of Overfitting}} & \multirow{2}{*}{$\mathbf{\beta}$}\\
    & 	&  	
    			& \textbf{MSE}		& \textbf{MED}		& $\mathbf{R^2}$
    					& \textbf{MED}		& \textbf{MED}		& $\mathbf{R^2}$ 	&		\\ \hline \hline

	Baseline	& Any 			& Any
				& 1.0000			& 0.9962			& 0.0000			
						& 1.0000			& 1.0000			& 1.0000			& --	\\ \hline \hline
	\multirow{6}{*}{Linear}
				& ANN			& Correct
				& \textbf{0.6153}	& \textbf{0.7590}	& \textbf{0.3701}
						& 36.5317			& 6.4187			& \textbf{2.6555}	& -- 	\\ \cline{3-10}
				& (2048)		& Shuffled
				& 1.1804			& 1.0641			& -0.1815
						& 49.7003			& 7.5766			& -5.3788	 		& --	\\ \cline{2-10}
				
				& Pixel			& Correct
				& 1.3172			& 1.0845			& -0.3251
						& 2.6191			& 1.6310			& -1.5199	 		& --	\\ \cline{3-10}
				& (1875)		& Shuffled
				& 1.5915			& 1.2170			& -0.5860
						& 2.5953			& 1.6424			& -0.6625			& --	\\ \cline{2-10}

				& Pixel			& Correct
				& 1.2073			& 1.0428			& -0.2120
						& \textbf{2.3360}	& \textbf{1.5433}	& -2.2664 			& --	\\ \cline{3-10}
				& (507)		& Shuffled
				& 1.5077			& 1.1880			& -0.5032
						& 2.3792			& 1.5735			& -0.7302 			& --	\\ \hline \hline
	\multirow{3}{*}{Lasso}
				& ANN (2048) 			& Correct
				& \textbf{0.5711}	& \textbf{0.7249}	& \textbf{0.4172}
						& 20.8883			& 4.8409			& 2.3302			& 0.01 	\\ \cline{2-10}
				
				& Pixel	(1875)		& Correct
				& 0.9183			& 0.9391			& 0.0788
						& \textbf{1.1320}	& 1.1371			& 2.3313 			& 0.2, 0.5	\\ \cline{2-10}

				& Pixel (507)			& Correct
				& 0.8946			& 0.9292			& 0.1015
						& 1.1677			& \textbf{1.1251}	& \textbf{2.2538}	& 0.05, 0.1	\\ \hline
						
\end{tabular}
\caption{Performance of the different regressors for different feature spaces and correct vs. shuffled targets on the four-dimensional space by \cite{Horst2016}. The best results for each combination of column and regressor are highlighted in boldface.}
\label{Hybrid:tab:NOUN_ex1_regression}
\end{table}

Table \ref{Hybrid:tab:NOUN_ex1_regression} shows the results obtained in our experiment, grouped by the regression algorithm, feature space, and target mapping used. We have also reported the observed degree of overfitting. It is calculated by dividing training set performance by test set performance. Perfect generalization would result in an amount of overfitting of one, whereas larger values represent the factor to which the regression is more successful on the training set than on the test set. Let us for now only consider the linear regression.

We first focus on the results obtained on the ANN-based feature set. As we can see, the linear regression is able to beat the baseline when trained on the correct targets. The overall approach therefore seems to be sound. However, we see strong overfitting tendencies, showing that there is still room for improvement. When trained on the shuffled targets, the linear regression completely fails to generalize to the test set. This shows that the correct mapping (having a semantic meaning) is easier to learn than an unstructured mapping. In other words, the semantic structure of the similarity space makes generalization possible.

Let us now consider the pixel-based feature spaces. For both of these spaces, we observe that  linear regression performs worse than the baseline. Moreover, we can see that learning the shuffled mapping results in even poorer performance than learning the correct mapping. Due to the overall poor performance, we do not observe very strong overfitting tendencies. Finally, when comparing the two pixel-based feature spaces, we observe that the linear regression tends to perform better on the low-dimensional feature space than on the high-dimensional one. However, these performance differences are relatively small.

Overall, ANN-based features seem to be much more useful for our mapping task than the simple pixel-based features, confirming our observations from Section \ref{NOUN}.\\

In order to further improve our results, we now varied the regularization factor $\beta$ of the lasso regressor for all feature spaces. 

For the ANN-based feature space, we are able to achieve a slight but consistent improvement by introducing a regularization term: Increasing $\beta$ causes poorer performance on the training set while yielding improvements on the test set. The best results on the test set are achieved for $\beta = 0.01$. If $\beta$ however becomes too large, then performance on the test set starts to decrease again -- for $\beta = 0.2$ we do not see any improvements over the vanilla linear regression any more. For $\beta \geq 5$, the lasso regression collapses and performs worse than the baseline. 

Although we are able to improve our performance slightly, the gap between training set performance and test set performance still remains quite high. It seems that the overfitting problem can be somewhat mitigated but not solved on our data set with the introduction of a simple regularization term.

When comparing our best results to the ones obtained by \cite{Sanders2018} who achieved values of $R^2 \approx 0.8$, we have to recognize that our approach performs considerably worse with $R^2 \approx 0.4$. However, the much smaller number of data points in our experiment makes our learning problem much harder than theirs. Even though we use data augmentation, the small number of different targets might put a hard limit on the quality of the results obtainable in this setting. Moreover, Sanders and Nosofsky retrained the whole neural network in their experiments whereas we limit ourselves to the features extracted by the pretrained network. As we are nevertheless able to clearly beat our baselines, we take these results as supporting the general approach.

For the pixel-based feature spaces we can also observe positive effects of regularization. For the large space, the best results on the test set are achieved for larger values of $\beta \in \{0.2, 0.5\}$. These results are however only slightly better than baseline performance. 
For the small pixel-based feature space, the optimal value of $\beta$ lies in $\{0.05, 0.1\}$, leading again to a test set performance comparable to the baseline. In case of the small pixel-based feature space, already relatively small degrees of regularization ($\beta \geq 1$) lead to a collapse of the model. 

Comparing the regularization results on the three feature spaces, we can conclude that regularization is indeed helpful, but only to a small degree. On the ANN-based feature space, we still observe a large amount of overfitting, and performance on the pixel-based feature spaces is still relatively close to the baseline. Looking at the optimal values of $\beta$, it seems like the lower-dimensional pixel-based feature space needs less regularization than its higher-dimensional counterpart. Presumably, this is caused by the smaller possibility for overfitting in the lower-dimensional feature space. Even though the larger pixel-based feature space and the ANN-based feature space have a similar dimensionality, the pixel-based feature space requires a larger degree of regularization for obtaining optimal performance, indicating that it is more prone to overfitting than the ANN-based feature space.

\subsection{Experiment 2: Comparing MDS Algorithms}
\label{ML:Experiment_2}

After having analyzed the soundness of our approach in experiment 1, we compare target spaces of the same dimensionality, but obtained from different MDS algorithms. More specifically, we compare the results from experiment 1 to results obtainable on the four-dimensional spaces created by both metric and nonmetric SMACOF in Section \ref{NOUN}.
Table \ref{tab:ex2_regression} shows the results obtained in our second experiment. 

\begin{table}[t]
  \centering
  \begin{tabular}{|c|c||c|c|c||c|c|c||c|}
    \hline
    \multirow{2}{*}{\textbf{Regressor}} & \multirow{2}{*}{\textbf{Target Space}} & \multicolumn{3}{|c||}{\textbf{Test Set Performance}} & \multicolumn{3}{|c||}{\textbf{Amount of Overfitting}} & \multirow{2}{*}{$\mathbf{\beta}$} \\
    	&		& \textbf{MSE}		& \textbf{MED}		& $\mathbf{R^2}$
    					& \textbf{MSE}		& \textbf{MED}		& $\mathbf{R^2}$ &	\\ \hline \hline
		& Horst and Hout
				& 1.0000			& 0.9962			& 0.0000
    					& 1.0000			& 1.0000			& 1.0000	& -- 	\\ \cline{2-9}
	Baseline
		& Metric SMACOF
				& 1.0000			& 0.9981			& 0.0000
    					& 1.0000			& 1.0000			& 1.0000	& -- 	\\ \cline{2-9}
		& Nonmetric SMACOF
				& 1.0000			& \textbf{0.9956}	& 0.0000
    					& 1.0000			& 1.0000			& 1.0000	& -- 	\\ \hline \hline

		& Horst and Hout
				& \textbf{0.6153}	& \textbf{0.7590}	& \textbf{0.3701}
    					& 36.5317			& 6.4187			& \textbf{2.6555}	& -- 	\\ \cline{2-9}
	Linear
		& Metric SMACOF
				& 0.6334			& 0.7697			& 0.3599
    					& \textbf{36.2300}	& 6.3909			& 2.7299	& --	\\ \cline{2-9}
		& Nonmetric SMACOF
				& 0.6225			& \textbf{0.7590}	& 0.3565
    					& 36.3861			& \textbf{6.3749}	& 2.7560	& --	\\ \hline \hline

		& Horst and Hout
				& \textbf{0.5711}	& \textbf{0.7249}	& \textbf{0.4172}	
    					& 20.8883			& 4.8409			& \textbf{2.3302}	& 0.01 	\\ 	\cline{2-9}
	Lasso
		& Metric SMACOF
				& 0.6160			& 0.7513			& 0.3766			
    					& 21.5633			& 3.7147			& 2.5789	& 0.01, 0.05 \\ \cline{2-9}
		& Nonmetric SMACOF
				& 0.5834			& 0.7284			& 0.3969			
    					& \textbf{11.9882}	& \textbf{3.6440}	& 2.3936	& 0.05	\\ \hline
		
 \end{tabular}
\caption{Comparison of the results obtainable on four-dimensional spaces created by different MDS algorithms. Best results in each column are highlighted for each of the regressors.}
\label{tab:ex2_regression}
\end{table}

In a first step, we can compare the different target spaces by taking a look at the behavior of the zero baseline in each of them. As we can see, the values for MSE and $R^2$ are identical for all of the different spaces. Only for the MED we can observe some slight variations, which can be explained by the slightly different arrangements of points in the different similarity spaces. 

As we can see from Table \ref{tab:ex2_regression}, the results for the linear regression on the different target spaces are comparable. This adds further support to our results from Section \ref{NOUN}: Also when considering the usage as target space for machine learning, metric MDS does not seem to have any advantage over nonmetric MDS.

Also for the lasso regressor we observed similar effects for all of the target spaces: A certain amount of regularization is helpful to improve test set performance, while too much emphasis on the regularization term causes both training and test set performance to collapse. Again, we still observe a large amount of overfitting even after using regularization.
Again, the results are comparable across the different target spaces. However, the optimal performance on the space obtained with metric SMACOF is consistently worse than the results obtained on the other two spaces. As the space by Horst and Hout is however also based on metric MDS, we cannot use this observation as an argument for nonmetric MDS.

\subsection{Experiment 3: Comparing Target Spaces of Different Size}
\label{ML:Experiment_3}

\begin{table}[t]
  \centering
  \begin{tabular}{|c|c||c|c|c||c|c|c||c|}
    \hline
    \multirow{2}{*}{\textbf{Regressor}}	& \multirow{2}{*}{$\mathbf{t}$}	& \multicolumn{3}{|c||}{\textbf{Test Set Performance}}	& 	\multicolumn{3}{|c||}{\textbf{Amount of Overfitting}} & \multirow{2}{*}{$\mathbf{\beta}$}\\
    	&	& \textbf{MSE}		& \textbf{MED}		& $\mathbf{R^2}$
    				& \textbf{MSE}		& \textbf{MED}		& $\mathbf{R^2}$ 	&	\\ \hline \hline
    \multirow{10}{*}{Baseline}
    	& 1	
    		& 1.0000			& \textbf{0.8665}	& 0.0000
    				& 1.0000			& 1.0000			& 1.0000			& -- \\ \cline{2-9}
    	& 2
    		& 1.0000			& 0.9581   			& 0.0000
    				& 1.0000			& 1.0000			& 1.0000			& -- \\ \cline{2-9}
    	& 3
    		& 1.0000			& 0.9848   			& 0.0000
    				& 1.0000			& 1.0000			& 1.0000			& -- \\ \cline{2-9}
    	& 4
    		& 1.0000			& 0.9956   			& 0.0000
    				& 1.0000			& 1.0000			& 1.0000			& -- \\ \cline{2-9}
    	& 5
    		& 1.0000			& 0.9965   			& 0.0000
    				& 1.0000			& 1.0000			& 1.0000			& -- \\ \cline{2-9}
    	& 6
    		& 1.0000			& 0.9973   			& 0.0000
    				& 1.0000			& 1.0000			& 1.0000			& -- \\ \cline{2-9}
    	& 7
    		& 1.0000			& 0.9978   			& 0.0000
    				& 1.0000			& 1.0000			& 1.0000			& -- \\ \cline{2-9}
    	& 8
    		& 1.0000			& 0.9980   			& 0.0000
    				& 1.0000			& 1.0000			& 1.0000			& -- \\ \cline{2-9}
    	& 9
    		& 1.0000			& 0.9982   			& 0.0000
    				& 1.0000			& 1.0000			& 1.0000			& -- \\ \cline{2-9}
    	& 10
    		& 1.0000			& 0.9984   			& 0.0000
    				& 1.0000			& 1.0000			& 1.0000			& -- \\ \hline \hline

    \multirow{10}{*}{Linear}
    	& 1
    		& 1.1547			& 0.9012   			& -0.1547
    				& 50.7798			& 7.6907			& -6.3153			& -- \\ \cline{2-9}
    	& 2
    		& \textbf{0.5087}	& \textbf{0.6468}	& \textbf{0.4869}
    				& \textbf{33.6837}	& \textbf{6.1353}	& \textbf{2.0226}	& -- \\ \cline{2-9}
    	& 3
    		& 0.5589			& 0.7022   			& 0.4380
    				& 35.0878			& 6.2391			& 2.2465			& -- \\ \cline{2-9}
    	& 4
    		& 0.6225			& 0.7590   			& 0.3565
    				& 36.3861			& 6.3749			& 2.7560			& -- \\ \cline{2-9}
    	& 5
    		& 0.6420			& 0.7770   			& 0.3512
    				& 37.2875			& 6.4258			& 2.7977			& -- \\ \cline{2-9}
    	& 6
    		& 0.6441			& 0.7800   			& 0.3396
    				& 37.1193			& 6.3781			& 2.8929			& -- \\ \cline{2-9}
    	& 7
    		& 0.6753			& 0.8026   			& 0.3193
    				& 38.0741			& 6.4568			& 3.0760			& -- \\ \cline{2-9}
    	& 8
    		& 0.6859			& 0.8127   			& 0.2996
    				& 38.1881			& 6.4699			& 3.2767 			& -- \\ \cline{2-9}
    	& 9
    		& 0.6878			& 0.8138   			& 0.2936
    				& 38.2118			& 6.4540			& 3.3441			& -- \\ \cline{2-9}
    	& 10
    		& 0.7136			& 0.8299   			& 0.2754
    				& 39.1195			& 6.5218			& 3.5637			& -- \\ \hline \hline

    \multirow{10}{*}{Lasso}
    	& 1
    		& 1.0306			& 0.8826			& -0.0306
    				& \textbf{1.0327}	& \textbf{1.0206}	& -0.0667			& 5, 10 \\ \cline{2-9}
    	& 2
    		& \textbf{0.4808}	& \textbf{0.6064}	& \textbf{0.5174}
    				& 21.1853			& 4.3094			& \textbf{1.8888}	& 0.01, 0.02 \\ \cline{2-9}
    	& 3
    		& 0.5274			& 0.6740			& 0.4715
    				& 17.1673			& 4.3433			& 2.0556			& 0.02 \\ \cline{2-9}
    	& 4
    		& 0.5834			& 0.7284			& 0.3969
    				& 11.9882			& 3.6440			& 2.3936			& 0.05 \\ \cline{2-9}
    	& 5
    		& 0.6229			& 0.7627			& 0.3720
    				& 25.6519			& 5.3451			& 2.6223			& 0.005 \\ \cline{2-9}
    	& 6
    		& 0.6333			& 0.7726			& 0.3498
    				& 30.5197			& 5.8073			& 2.7982			& 0.002 \\ \cline{2-9}
    	& 7
    		& 0.6489			& 0.7813			& 0.3418
    				& 16.5268			& 4.2493			& 2.8098			& 0.02 \\ \cline{2-9}
    	& 8
    		& 0.6729			& 0.8037			& 0.3117
    				& 30.6533			& 5.8197			& 3.1370			& 0.002 \\ \cline{2-9}
    	& 9
    		& 0.6721			& 0.8019			& 0.3092
    				& 20.4216			& 4.7331			& 3.1257			& 0.01 \\ \cline{2-9}
    	& 10
    		& 0.7080			& 0.8261			& 0.2810
    				& 34.6693			& 6.1649			& 3.4854			& 0.001 \\ \hline
    
   \end{tabular}
\caption{Performance of the zero baseline, the linear regression, and the lasso regression  on target spaces of different dimensionality $t$ derived with nonmetric SMACOF, along with the relative amount of overfitting. Best values for each column are highlighted for each of the regressors.}
\label{tab:ex3_regression}
\end{table}

In our third and final experiment in this study, we vary the number of dimensions in the target space. More specifically, we consider similarity spaces with one to ten dimensions that have been created by nonmetric SMACOF.

\begin{figure}[t]
	\centering
  	\includegraphics[width = 1.0\columnwidth]{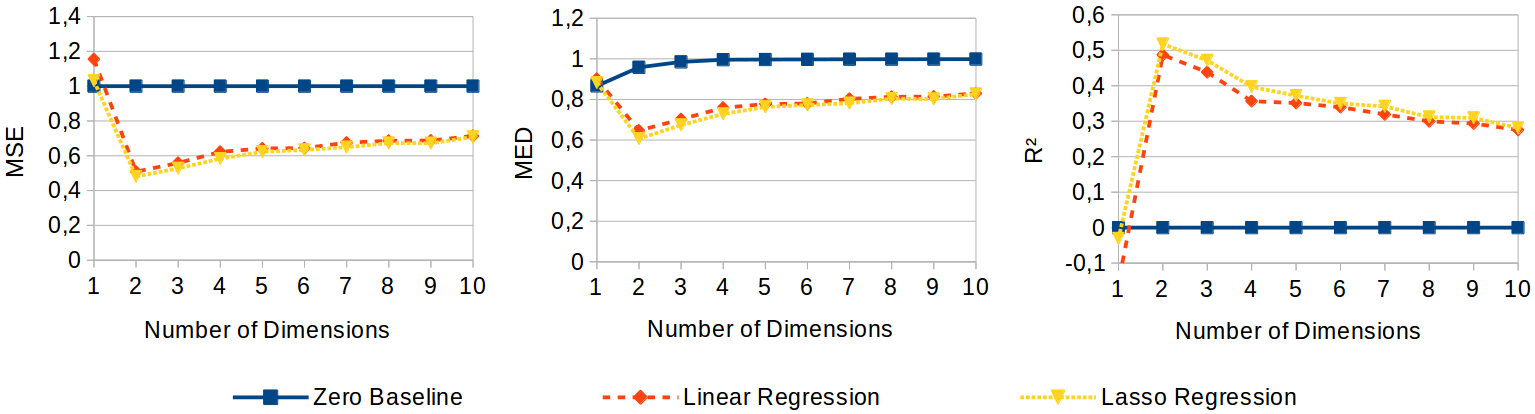}
	\caption{Visualization of the regression results for MSE, MED, and $R^2$ as a function of the number of dimensions.}
	\label{fig:ex3_regression}
\end{figure}

Table \ref{tab:ex3_regression} displays the results obtained in our third experiment and Figure \ref{fig:ex3_regression} provides a graphical illustration. When looking at the zero baseline, we observe that the mean Euclidean distance tends to grow with an increasing number of dimensions, with an asymptote of one. This indicates that in higher-dimensional spaces, the points seem to lie closer to the surface of a unit hypershpere around the origin. For both MSE and $R^2$ we do not observe any differences between the target spaces.\\

Let us now look at the results of the linear regression. It seems that for all the evaluation metrics, a two-dimensional target space yields the best result. With an increasing number of dimensions in the target space, performance tends to decrease. We can also observe that the amount of overfitting is optimal for a two-dimensional space and tends to increase with an increasing number of dimensions. A notable exception is the one-dimensional space which suffers strongly from overfitting and whose performance with respect to all three evaluation metrics is clearly worse than the zero baseline.

The optimal performance of a lasso regressor on the different target spaces when trained on the ANN-based features yields similar results: For all target spaces we made again the observation the regularization can help to improve performance but that too much regularization decreases performance. Again, we can only counteract a relatively small amount of the observed overfitting. As we can see in Table \ref{tab:ex3_regression}, again a two-dimensional space yields the best results. 

Taken together, the results of our third experiment show that a higher-dimensional target space makes the regression problem more difficult, but that a one-dimensional target space does not contain enough semantic structure for a successful mapping. It seems that a two-dimensional space is in our case the optimal trade-off. However, even the performance of the optimal regressor on this space is far from satisfactory, urging for further research.

\section{Conclusions}
\label{Conclusions}

The contributions of this paper are twofold:

In our first study, we investigated whether the dissimilarity ratings obtained throught SpAM are ratio scaled by applying both metric MDS (which assumes a ratio scale) and nonmetric MDS (which only assumes an ordinal scale). Both MDS variants produced comparable results -- it thus seems that assuming a ratio scale is neither beneficial nor harmful. We therefore recommend to use nonmetric MDS as its underlying assumptions are weaker. Future studies on other data sets obtained through SpAM should seek to confirm or contradict our results.

In our second study, we analyzed whether learning a mapping from raw images to points in a psychological similarity space is possible. Our results showed that using the activations of a pretrained ANN as features for a regression task seems to work in principle. However, we observed very strong overfitting tendencies in our experiments. Furthermore, the overall performance level we were able to achieve is still far from satisfactory. The results by \cite{Sanders2018} however show that larger amounts of training data can alleviate these problems. Future work in this area should focus on improvements in performance and robustness of this approach.

As follow-up work, we are currently conducting a study on a data set of shapes, where we plan to apply more sophisticated machine learning methods in order to counteract the observed overfitting tendencies.

\bibliographystyle{apalike}
\bibliography{/home/lbechberger/Documents/Papers/jabref.bib}{}

\begin{thebibliography}{}

\bibitem[Bechberger, 2019]{Bechberger2019GitHubPsy}
Bechberger, L. (2019).
\newblock {lbechberger/LearningPsychologicalSpaces: Study on Multidimensional
  Scaling and Neural Networks on the NOUN Dataset}.

\bibitem[Bechberger and Kypridemou, 2018]{Bechberger2018AIC}
Bechberger, L. and Kypridemou, E. (2018).
\newblock {Mapping Images to Psychological Similarity Spaces Using Neural
  Networks}.
\newblock In {\em Proceedings of the 6th International Workshop on Artificial
  Intelligence and Cognition}.

\bibitem[Borg and Groenen, 2005]{Borg2005}
Borg, I. and Groenen, J.~F. (2005).
\newblock {\em {Modern Multidimensional Scaling: Theory and Applications}}.
\newblock Springer Series in Statistics. Springer-Verlag New York, 2 edition.

\bibitem[de~Leeuw, 1977]{DeLeeuw1977}
de~Leeuw, J. (1977).
\newblock {\em Recent Development in Statistics}, chapter {Applications of
  Convex Analysis to Multidimensional Scaling}, pages 133--146.
\newblock North Holland Publishing.

\bibitem[G{\"a}rdenfors, 2000]{Gardenfors2000}
G{\"a}rdenfors, P. (2000).
\newblock {\em {C}onceptual {S}paces: {T}he {G}eometry of {T}hought}.
\newblock {M}{I}{T} press.

\bibitem[Goldstone, 1994]{Goldstone1994}
Goldstone, R. (1994).
\newblock {A}n {E}fficient {M}ethod for {O}btaining {S}imilarity {D}ata.
\newblock {\em Behavior Research Methods, Instruments, \& Computers},
  26(4):381--386.

\bibitem[Horst and Hout, 2016]{Horst2016}
Horst, J.~S. and Hout, M.~C. (2016).
\newblock {The Novel Object and Unusual Name (NOUN) Database: A Collection of
  Novel Images for Use in Experimental Research}.
\newblock {\em Behavior Research Methods}, 48(4):1393--1409.

\bibitem[Hout et~al., 2014]{Hout2014}
Hout, M.~C., Goldinger, S.~D., and Brady, K.~J. (2014).
\newblock {MM-MDS: A Multidimensional Scaling Database with Similarity Ratings
  for 240 Object Categories from the Massive Memory Picture Database}.
\newblock {\em PLOS ONE}, 9(11):1--11.

\bibitem[Hout et~al., 2013]{Hout2013}
Hout, M.~C., Goldinger, S.~D., and Ferguson, R.~W. (2013).
\newblock {T}he {V}ersatility of {S}p{AM}: {A} {F}ast, {E}fficient, {S}patial
  {M}ethod of {D}ata {C}ollection for {M}ultidimensional {S}caling.
\newblock {\em Journal of Experimental Psychology: General}, 142(1):256.

\bibitem[Kruskal, 1964a]{Kruskal1964}
Kruskal, J.~B. (1964a).
\newblock {Multidimensional Scaling by Optimizing Goodness of Fit to a
  Nonmetric Hypothesis}.
\newblock {\em Psychometrika}, 29(1):1--27.

\bibitem[Kruskal, 1964b]{Kruskal1964a}
Kruskal, J.~B. (1964b).
\newblock {Nonmetric Multidimensional Scaling: A Numerical Method}.
\newblock {\em Psychometrika}, 29(2):115--129.

\bibitem[Pearson, 1895]{Pearson1895}
Pearson, K. (1895).
\newblock {Note on Regression and Inheritance in the Case of Two Parents}.
\newblock {\em Proceedings of the Royal Society of London},
  58(347-352):240--242.

\bibitem[Peterson et~al., 2017]{Peterson2017}
Peterson, J.~C., Abbott, J.~T., and Griffiths, T.~L. (2017).
\newblock {Adapting Deep Network Features to Capture Psychological
  Representations: An Abridged Report}.
\newblock In {\em Proceedings of the Twenty-Sixth International Joint
  Conference on Artificial Intelligence, {IJCAI-17}}, pages 4934--4938.

\bibitem[Peterson et~al., 2018]{Peterson2018}
Peterson, J.~C., Abbott, J.~T., and Griffiths, T.~L. (2018).
\newblock {Evaluating (and Improving) the Correspondence Between Deep Neural
  Networks and Human Representations}.
\newblock {\em Cognitive Science}, 42(8):2648--2669.

\bibitem[Sanders and Nosofsky, 2018]{Sanders2018}
Sanders, C.~A. and Nosofsky, R.~M. (2018).
\newblock {Using Deep-Learning Representations of Complex Natural Stimuli as
  Input to Psychological Models of Classification}.
\newblock In {\em Proceedings of the 2018 Conference of the Cognitive Science
  Society, Madison.}

\bibitem[Spearman, 1904]{Spearman1904}
Spearman, C. (1904).
\newblock {The Proof and Measurement of Association between Two Things}.
\newblock {\em The American Journal of Psychology}, 15(1):72--101.

\bibitem[Szegedy et~al., 2016]{Szegedy2016}
Szegedy, C., Vanhoucke, V., Ioffe, S., Shlens, J., and Wojna, Z. (2016).
\newblock {Rethinking the Inception Architecture for Computer Vision}.
\newblock In {\em Proceedings of the IEEE Conference on Computer Vision and
  Pattern Recognition}, pages 2818--2826.

\bibitem[Wickelmaier, 2003]{Wickelmaier2003}
Wickelmaier, F. (2003).
\newblock {An Introduction to MDS}.
\newblock {\em Sound Quality Research Unit, Aalborg University, Denmark},
  46(5).

\end{thebibliography}

\end{document}